\documentclass[10pt,twocolumn,letterpaper]{article}

\usepackage{cvpr}              
\usepackage{array}
\usepackage[accsupp]{axessibility}

\usepackage[dvipsnames]{xcolor}

\definecolor{cvprblue}{rgb}{0.21,0.49,0.74}
\usepackage[pagebackref,breaklinks,colorlinks,citecolor=cvprblue]{hyperref}

\def\paperID{10967} 
\def\confName{CVPR}
\def\confYear{2024}

\title{TRIP: Temporal Residual Learning with Image Noise Prior\\ 
	   for Image-to-Video Diffusion Models\thanks{{\small This work was performed at HiDream.ai.}}}

\author{\normalsize Zhongwei Zhang$^{\dag}$, Fuchen Long$^{\S}$, Yingwei Pan$^{\S}$, Zhaofan Qiu$^{\S}$, Ting Yao$^{\S}$, Yang Cao$^{\dag}$  and Tao Mei$^{\S}$\\
	$^{\dag}$\normalsize University of Science and Technology of China, Hefei, China \\ 
	$^{\S}$\normalsize HiDream.ai Inc. \\
	{\tt\small\ zhwzhang@mail.ustc.edu.cn}, {\tt\small\{longfuchen, pandy, qiuzhaofan, tiyao\}@hidream.ai} \\
	{\tt\small\ forrest@ustc.edu.cn}, {\tt\small\ tmei@hidream.ai} \\
}
\begin{document}
\maketitle

\begin{abstract}
Recent advances in text-to-video generation have demonstrated the utility of powerful diffusion models. Nevertheless, the problem is not trivial when shaping diffusion models to animate static image (i.e., image-to-video generation). The difficulty originates from the aspect that the diffusion process of subsequent animated frames should not only preserve the faithful alignment with the given image but also pursue temporal coherence among adjacent frames. To alleviate this, we present TRIP, a new recipe of image-to-video diffusion paradigm that pivots on image noise prior derived from static image to jointly trigger inter-frame relational reasoning and ease the coherent temporal modeling via temporal residual learning. Technically, the image noise prior is first attained through one-step backward diffusion process based on both static image and noised video latent codes. Next, TRIP executes a residual-like dual-path scheme for noise prediction: 1) a shortcut path that directly takes image noise prior as the reference noise of each frame to amplify the alignment between the first frame and subsequent frames; 2) a residual path that employs 3D-UNet over noised video and static image latent codes to enable inter-frame relational reasoning, thereby easing the learning of the residual noise for each frame. Furthermore, both reference and residual noise of each frame are dynamically merged via attention mechanism for final video generation. Extensive experiments on WebVid-10M, DTDB and MSR-VTT datasets demonstrate the effectiveness of our TRIP for image-to-video generation. Please see our project page at \url{https://trip-i2v.github.io/TRIP/}.
\end{abstract}

\begin{figure}
	\centering
	\includegraphics[width=0.95\linewidth]{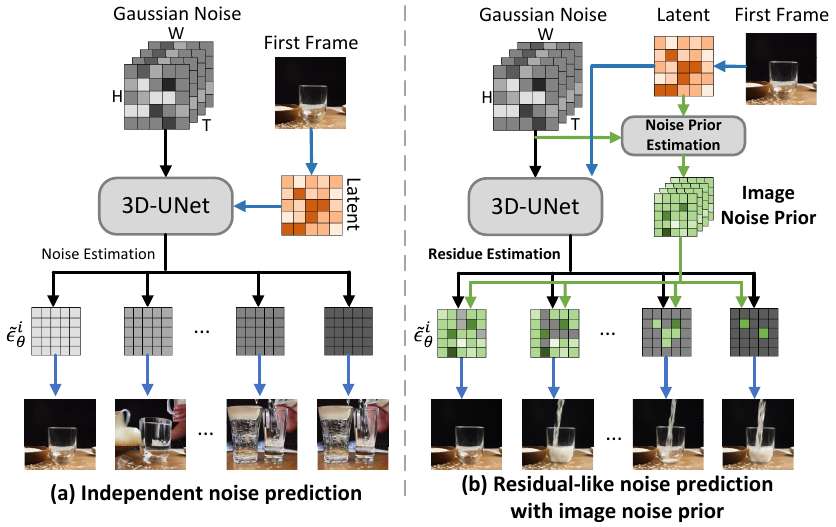}
	\vspace{-0.15in}
	\caption{Comparison between (a) independent noise prediction and (b) our residual-like noise prediction with image noise prior in I2V diffusion models.}
	\label{fig:intro}
	\vspace{-0.23in}
\end{figure}

\section{Introduction}
\label{sec:intro}

In recent years, deep generative models have demonstrated impressive capabilities to create high-quality visual content. In between, Generative Adversarial Networks (GANs) \cite{2014gan,2017cyclegan} and diffusion models \cite{2020ddpm,2021beat-gans} have brought forward milestone improvement for a series of generative tasks in computer vision field, e.g., text-to-image generation \cite{2021glide,2022ldm}, image editing \cite{2023instructpix2pix,2022prompt}, and text-to-video generation \cite{2023align,2022imagen,2022vdm}. In this work, we are interested in image-to-video generation (I2V), i.e., animating a static image by endowing it with motion dynamics, which has great potential real-world applications (e.g., entertainment and creative media). Conventional I2V techniques \cite{2022simulating,2022controllable,2021stochastic} commonly target for generating video from a natural scene image with random or very coarse motion dynamics (such as clouds moving or fluid flowing). In contrast, our focus is a more challenging text-driven I2V scenario: given the input static image and text prompt, the animated frames should not only faithfully align with the given first frame, but also be temporally coherent and semantically matched with text prompt.

Several recent pioneering practices \cite{2023videocomposer, 2023seer} for I2V is to remould the typical latent diffusion model of text-to-video generation by directly leveraging the given static image as additional condition during diffusion process.
Figure \ref{fig:intro} (a) illustrates such conventional noise prediction strategy in latent diffusion model. 
Concretely, by taking the given static image as the first frame, 2D Variational Auto-encoder (VAE) is first employed to encode the first frame as image latent code. This encoded image latent code is further concatenated with noised video latent code (i.e., a sequence of Gaussian noise) to predict the backward diffusion noise of each subsequent frame via a learnable 3D-UNet.
Nevertheless, such independent noise prediction strategy of each frame leaves the inherent relation between the given image and each subsequent frame under-exploited, and is lacking in efficacy of modeling temporal coherence among adjacent frames.
As observed in I2V results of Figure \ref{fig:intro} (a), both the foreground content (a glass) and the background content (e.g., a table) in the synthesized second frame are completely different from the ones in the given first frame.
To alleviate this issue, our work shapes a new way to formulate the noise prediction in I2V diffusion model as temporal residual learning on the basis of amplified guidance of given image (i.e., image noise prior). As illustrated in Figure \ref{fig:intro} (b), our unique design is to integrate the typical noise prediction with an additional shortcut path that directly estimates the backward diffusion noise of each frame by solely referring image noise prior. Note that this image noise prior is calculated based on both input image and noised video latent codes, which explicitly models the inherent correlation between the given first frame and each subsequent frame. Meanwhile, such residual-like scheme re-shapes the typical noise prediction via 3D-UNet as residual noise prediction to ease temporal modeling among adjacent frames, thereby leading to more temporally coherent I2V results.

By materializing the idea of executing image-conditioned noise prediction in a residual manner, we present a novel diffusion model, namely Temporal Residual learning with Image noise Prior (TRIP), to facilitate I2V.
Specifically, given the input noised video latent code and the corresponding static image latent code, TRIP performs residual-like noise prediction along two pathways. One is the shortcut path that first achieves image noise prior based on static image and noised video latent codes through one-step backward diffusion process, which is directly regarded as the reference noise of each frame. The other is the residual path which concatenates static image and noised video latent codes along temporal dimension and feeds them into 3D-UNet to learn the residual noise of each frame. Eventually, a Transformer-based temporal noise fusion module is leveraged to dynamically fuse the reference and residual noises of each frame, yielding high-fidelity video that faithfully aligns with given image.

\section{Related Work}
\label{sec:relatd_work}

\textbf{Text-to-Video Diffusion Models.} 
The great success of diffusion models \cite{2021glide,2022ldm} to generate fancy images based on text prompts has been witnessed in recent years.  
The foundation of the basic generative diffusion models further encourages the development of customized image synthesis, such as image editing \cite{2023zero,2022prompt,2023instructpix2pix} and personalized image generation \cite{2022image,2023dreambooth,2023multi,2022dreamartist,2023adding,2023t2i-adapter}.
Inspired by the impressive results of diffusion models on image generation, a series of text-to-video (T2V) diffusion models \cite{2023align,2022flexible,2022vdm,2023text2video-zero,2023videofusion,2022diffusion,2022imagen,2022make-a-video,2023modelscope} start to emerge.
VDM \cite{2022vdm} is one of the earlier works that remoulds text-to-image (T2I) 2D-UNet architecture with temporal attention for text-to-video synthesis.
Later in \cite{2023align} and \cite{2023animatediff}, the temporal modules (e.g., temporal convolution and self-attention) in diffusion models are solely optimized to emphasize motion learning.
To further alleviate the burden on temporal modeling, Ge \emph{et al.} \cite{2023pyoco} present a mixed video noise model under the assumption that each frame should correspond to a shared common noise and its own independent noise.
Similar recipe is also adopted by VideoFusion \cite{2023videofusion} which estimates the basic noise and residual noise via two image diffusion models for video denoising.
Other works go one step further by applying diffusion models for customized video synthesis, such as video editing \cite{2023videdit,2023tokenflow,2023stablevideo,2023codef,2023magicprop} and personalized video generation \cite{2023animatediff,2023animate-a-story,2023magicavatar}.
In this work, we choose the basic T2V diffusion model with 3D-UNet \cite{2023modelscope} as our backbone for image-to-video generation task.

\textbf{Image-to-Video Generation.} 
Image-to-Video (I2V) generation is one kind of conditional video synthesis, where the core condition is a given static image. 
According to the availability of motion cues, I2V approaches can be grouped into two categories: stochastic generation (solely using input image as condition) \cite{2017stochastic,2018flow,2018learning,2021animating} and conditional generation (using image and other types of conditions) \cite{2022controllable,2022latent,2023conditional,2023dreampose,videodrafter}.
Early attempts of I2V generation mostly belong to the stochastic models which usually focus on the short motion of fluid elements (e.g., water flow) \cite{2021animating, 2022simulating} or human poses \cite{2023leo} for landscape or human video generation.
To further enhance the controllability of motion modeling, the conditional models exploit more types of conditions like text prompts \cite{2023seer,2021stochastic} in the procedure of video generation.
Recent advances start to explore diffusion models \cite{2023videocomposer,2023dragnvwa} for conditional I2V task.
One pioneering work (VideoComposer \cite{2023videocomposer}) flexibly composes a video with additional conditions of texts, optical flow or depth mapping sequence for I2V generation.
Nevertheless, VideoComposer independently predicts the noise of each frame and leaves the inherent relation between input image and other frames under-exploited, thereby resulting in temporal inconsistency among frames.

\begin{figure*}
	\centering
	\includegraphics[width=0.97\linewidth]{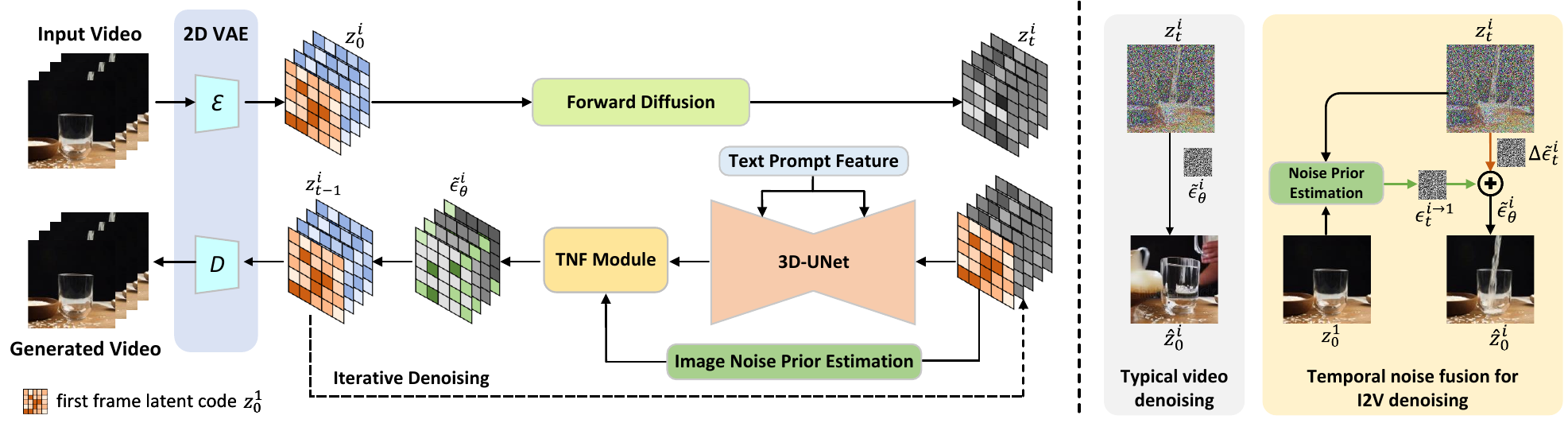}
	\vspace{-0.1in}
	\caption{An overview of (a) our Temporal Residual learning with Image noise Prior (TRIP), and (b) the conceptual comparison between typical frame-wise denoising and our temporal noise fusion. During training, the input video clip is encoded into a sequence of image latent codes via a pre-trained 2D VAE. The latent code of first frame is then concatenated with noised video latent code as the input of the 3D-UNet.
	Then, TRIP performs a residual-like noise prediction along two pathways, i.e., shortcut path and residual path.
	The shortcut path computes the image noise prior as the reference noise based on the first frame latent codes and the noised video latent codes through a one-step backward diffusion. 
	The residual path leverages 3D-UNet to predict the residual noise with reference to the image noise prior. 
    By dynamically merging the reference and residual noise via a Transformer-based Temporal Noise Fusion (TNF) module, the target noise vector for video denoising is attained. Both 3D-UNet and TNF module are jointly optimized with the typical MSE loss for I2V generation.}
	\label{fig:pipeline}
	\vspace{-0.1in}
\end{figure*}

\textbf{Deep Residual Learning.}
The effectiveness of learning residual components with additional shortcut connections in deep neural networks \cite{resnet,densenet,2018r2p1d} has been verified in various vision tasks. 
Besides typical convolution-based structure, recent advances \cite{ViT,Vaswani:NIPS17} also integrate the shortcut connection into the emerging transformer-based structure \cite{LiPAMI22,long2022dynamic,long2022sifa,wavevit2022,chen2023anchor}.
Intuitively, learning residual part with reference to the optimal identity mapping will ease the network optimization.   
In this work, we capitalize on such principle and formulate noise prediction in I2V diffusion as temporal residual learning on the basis of the image noise prior derived from static image to enhance temporal coherence.

In summary, our work designs a novel image-to-video diffusion paradigm with residual-like noise prediction. 
The proposed TRIP contributes by studying not only how to excavate the prior knowledge of given image as noise prior to guide temporal modeling, and how to strengthen temporal consistency conditioned on such image noise prior.


\section{Our Approach}
In this section, we present our Temporal Residual learning with Image noise Prior (TRIP) for I2V generation.
Figure \ref{fig:pipeline} illustrates an overview of our TRIP. 
Given a video clip at training, TRIP first extracts the per-frame latent code via a pre-trained VAE and groups them as a video latent code. 
The image latent code of the first frame (i.e., the given static image) is regarded as additional condition, which is further concatenated with the noised video latent code along temporal dimension as the input of the 3D-UNet.
Next, TRIP executes a residual-like noise prediction along two pathways (i.e., shortcut path and residual path), aiming to trigger inter-frame relational reasoning.
In shortcut path, the image noise prior of each frame is calculated based on the correlation between the first frame and subsequent frames through one-step backward diffusion process.
In residual path, the residual noise is estimated by employing 3D-UNet over the concatenated noised video latent code plus text prompt.
Finally, a transformer-based Temporal Noise Fusion (TNF) module dynamically merges the image noise prior and the residual noise as the predicted noise for video generation.

\subsection{Preliminaries: Video Diffusion Model}
T2V diffusion models \cite{2022vdm,2023align} are usually constructed by remoulding T2I diffusion models \cite{2022hierarchical,2021glide}. 
A 3D-UNet with text prompt condition is commonly learnt to denoise a randomly sampled noise sequence in Gaussian distribution for video synthesis.
Specifically, given an input video clip with $N$ frames, a pre-trained 2D VAE encoder $\mathcal{E}(\cdot)$ first extracts latent code of each frame to constitute a latent code sequence $\{z_0^i\}_{i=1}^{N}$. 
Then, this latent sequence is grouped via concatenation along temporal dimension as a video latent code $z_0$.
Next, the Gaussian noise is gradually added to $z_0$ through forward diffusion procedure (length: $T$, noise scheduler: $\{\beta_t\}_{t=1}^{T}$). The noised latent code $z_t$ at each time step $t$ is thus formulated as:
\begin{equation}\label{eq:1}
	z_t = \sqrt{\bar{\alpha}_t}z_0 + \sqrt{1-\bar{\alpha}_t}\epsilon,\, \epsilon \sim \mathcal{N}(0, I), 
\end{equation}
where $\bar{\alpha}_t = \prod_{i=1}^{t}\alpha_t,\, \alpha_t = 1-\beta_t$ and $\epsilon$ is the adding noise vector sampled from standard normal distribution $\mathcal{N}(0, I)$. 3D-UNet with parameter $\theta$ aims to estimate the noise $\epsilon$ based on the multiple inputs (i.e., noised video latent code $z_t$, time step $t$, and text prompt feature $c$ extracted by CLIP \cite{2021clip}). 
The Mean Square Error (MSE) loss is leveraged as the final objective $L$:
\begin{equation}\label{eq:2}
	\mathcal{L} = \mathbb{E}_{\epsilon\sim\mathcal{N}(0,I),t,c} [\Vert\epsilon-\epsilon_{\theta}(z_t,t,c)\Vert^2].
\end{equation}
In the inference stage, the video latent code $\hat{z}_0$ is iteratively denoised by estimating noise $\epsilon_{\theta}$ in each time step via 3D-UNet, and the following 2D VAE decoder $\mathcal{D}(\cdot)$ reconstructs each frame based on the video latent code $\hat{z}_0$.

\subsection{Image Noise Prior}
Different from typical T2V diffusion model, I2V generation model goes one step further by additionally emphasizing the faithful alignment between the given first frame and the subsequent frames. Most existing I2V techniques \cite{2023videocomposer,2023seer} control the video synthesis by concatenating the image latent code of first frame with the noised video latent code along channel/temporal dimension. 
In this way, the visual information of first frame would be propagated across all frames via the temporal modules (e.g., temporal convolution and self-attention).
Nevertheless, these solutions only rely on the temporal modules to calibrate video synthesis and leave the inherent relation between given image and each subsequent frame under-exploited, which is lacking in efficacy of temporal coherence modeling.
To alleviate this issue, we formulate the typical noise prediction in I2V generation as temporal residual learning with reference to an image noise prior, targeting for amplifying the alignment between synthesized frames and first static image. 

Formally, given the image latent code $z_0^1$ of first frame, we concatenate it with the noised video latent code sequence $\{z_t^i\}_{i=1}^{N}$ ($t$ denotes time steps in forward diffusion) along temporal dimension, yielding the conditional input $\{z_0^1, z_t^1, z_t^2, ..., z_t^N\}$ for 3D-UNet.
Meanwhile, we attempt to excavate the image noise prior to reflect the correlation between the first frame latent code $z_0^1$ and $i$-th noised frame latent code $z_t^i$.
According to Eq. (\ref{eq:1}), we can reconstruct the $i$-th frame latent code $z_0^i$ in one-step backward diffusion process as follows:
\begin{equation}\label{eq:3}
	z_0^i = \frac{z_t^i - \sqrt{1-\bar{\alpha}_t}\epsilon_t^i}{\sqrt{\bar{\alpha}_t}},
\end{equation}
where $\epsilon_t^i$ denotes Gaussian noise which is added to the $i$-th frame latent code. 
Considering the basic assumption in I2V that all frames in a short video clip are inherently correlated to the first frame, the $i$-th frame latent code $z_0^i$ can be naturally formulated as the combination of the first frame latent code $z_0^1$ and a residual item $\Delta z^i$ as:
\begin{equation}\label{eq:4}
	z_0^i = z_0^1 + \Delta z^i.
\end{equation}
Next, we construct a variable $C_t^i$ by adding a scale ratio to the residual item $\Delta z^i$ as follows:
\begin{equation}\label{eq:5}
	C_t^i = \frac{\sqrt{\bar{\alpha}_t}\Delta z^i}{\sqrt{1-\bar{\alpha}_t}}.
\end{equation}
Thus Eq. (\ref{eq:5}) can be re-written as follows:
\begin{equation}\label{eq:6}
	\Delta z^i = \frac{\sqrt{1-\bar{\alpha}_t}C_t^{i}}{\sqrt{\bar{\alpha}_t}}.
\end{equation}
After integrating Eq. (\ref{eq:3}) and Eq. (\ref{eq:6}) into Eq. (\ref{eq:4}), the first frame latent code $z_0^1$ is denoted as:
\begin{align}\label{eq:7} 
	z_0^1 & = z_0^i - \Delta z^i = \frac{z_t^i-\sqrt{1-\bar{\alpha}_t}\epsilon_t^i}{\sqrt{\bar{\alpha}_t}} - \frac{ \sqrt{1-\bar{\alpha}_t}C_t^{i}}{\sqrt{\bar{\alpha}_t}} \notag \\
	& = \frac{z_t^i- \sqrt{1-\bar{\alpha}_t}(\epsilon_t^i + C_t^{i})}{\sqrt{\bar{\alpha}_t}} = \frac{z_t^i- \sqrt{1-\bar{\alpha}_t}\epsilon_t^{i\to1}}{\sqrt{\bar{\alpha}_t}}, 
\end{align}
where $\epsilon_t^{i\to1}$ is defined as the image noise prior which can be interpreted as the relationship between the first frame latent code $z_0^1$ and $i$-th noised frame latent code $z_t^i$.
The image noise prior $\epsilon_t^{i\to1}$ is thus measured as:
\begin{equation}\label{eq:8}
	\epsilon_t^{i\to1} = \frac{z_t^i-\sqrt{\bar{\alpha}_t}z_0^1}{\sqrt{1-\bar{\alpha}_t}}.
\end{equation}
According to Eq. (\ref{eq:8}), the latent code $z_0^1$ of first frame can be directly reconstructed through the one-step backward diffusion by using the image noise prior $\epsilon_t^{i\to1}$ and the noised latent code $z_t^i$ of $i$-th frame.
In this way, such image noise prior $\epsilon_t^{i\to1}$ acts as a reference noise for Gaussian noise $\epsilon_t^i$ which is added in $i$-th frame latent code, when $z_0^i$ is close to $z_0^1$ (i.e., $i$-th frame is similar to the given first frame).

\subsection{Temporal Residual Learning}
Recall that the image noise prior learnt in the shortcut pathway facilitates temporal modeling to amplify the alignment between the first frame and subsequent frames. 
To further strengthen temporal coherence among all adjacent frames, a residual pathway is designed to estimate the residual noise of each frame through inter-frame relational reasoning.
Inspired by exploring knowledge prior in pre-trained 3D-UNet for video editing \cite{2023glv}, we propose to fine-tune 3D-UNet to estimate the residual noise of each frame.
Technically, the estimated noise $\tilde{\epsilon}_t^i$ of $i$-th frame is formulated as the combination of the reference noise $\epsilon_t^{i\to1}$ (i.e., image noise prior) and the estimated residual noise $\Delta \tilde{\epsilon}_t^i$: 
\begin{equation}\label{eq:9}
	\tilde{\epsilon}_t^i = \lambda^i\epsilon_t^{i\to1}+(1-\lambda^i)\Delta \tilde{\epsilon}_t^i,
\end{equation} 
where $\lambda^i$ is a trade-off parameter. $\Delta \tilde{\epsilon}_t^i$ is learnt by 3D-UNet (parameter: $\theta$), and $\epsilon_t^{i\to1}$ can be computed by Eq. (\ref{eq:8}) on the fly. 
Finally, the I2V denoising diffusion objective $\tilde{L}$ is calculated as follows:
\begin{equation} \label{eq:10}
	 \tilde{\mathcal{L}} = \mathbb{E}_{\epsilon\sim\mathcal{N}(0,I),t,c,i} [\Vert\epsilon_t^i-\tilde{\epsilon}_t^i\Vert^2].
\end{equation}
Note that the computation of Eq. (\ref{eq:9}) is operated through the Temporal Noise Fusion (TNF) module that estimates the backward diffusion noise of each frame.
Moreover, since the temporal correlation between $i$-th frame and the first frame will decrease when increasing frame index $i$ with enlarged timespan, we shape the trade-off parameter $\lambda^i$ as a linear decay parameter with respect to frame index.

\begin{figure}
	\centering
  \vspace{-0.1in}
	\includegraphics[width=0.80\linewidth]{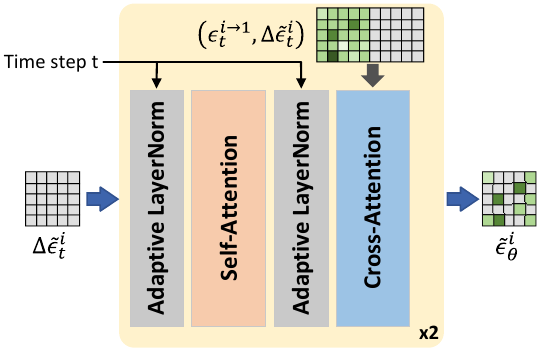}
	\vspace{-0.1in}
	\caption{Detailed structure of Temporal Noise Fusion module.}
	\label{fig:transformer}
	\vspace{-0.2in}
\end{figure}

\subsection{Temporal Noise Fusion Module}
Our residual-like dual-path noise prediction explores the image noise prior as a reference to facilitate temporal modeling among adjacent frames. 
However, the handcrafted design of TNF module with simple linear fusion (i.e., Eq. (\ref{eq:9})) requires a careful tuning of hyper-parameter $\lambda^i$, leading to a sub-optimal solution.
Instead, we devise a new Transformer-based temporal noise fusion module to dynamically fuse reference and residual noises, aiming to further exploit the relations in between and boost noise fusion.

Figure \ref{fig:transformer} depicts the structure of Temporal Noise Fusion module. 
Specifically, given the estimated residual noise $\Delta \tilde{\epsilon}_t^i$, we first leverage the adaptive layer norm operations \cite{2019adanorm} which is modulated by the time step $t$ for spatial feature normalization.
After that, one self-attention layer is employed for feature enhancement, followed by the same adaptive layer normalization operation.
Next, by taking the in-process intermedia feature as query and the concatenated feature of [$\epsilon_t^{i\to1}$, $\Delta \tilde{\epsilon}_t^i$] as key/value, TNF module achieves the final backward diffusion noise $\tilde{\epsilon}_\theta(z_t, t, c, i)$ through a cross-attention layer:
\begin{equation}\label{eq:11}
\tilde{\epsilon}_\theta(z_t, t, c, i) = \tilde{\epsilon}_t^i = \varphi(\Delta \tilde{\epsilon}_t^i, \epsilon^{i\to1}, t),
\end{equation}   
where $\varphi(\cdot)$ denotes the operation of TNF module.
Compared to the simple linear fusion strategy for noise estimation, our Transformer-based TNF module sidesteps the hyper-parameter tuning and provides an elegant alternative to dynamically merge the reference and residual noises of each frame to yield high-fidelity videos.

\section{Experiments}

\subsection{Experimental Settings}

\textbf{Datasets.} 
We empirically verify the merit of our TRIP model on WebVid-10M \cite{2021Webvid}, DTDB \cite{2018DTDB} and MSR-VTT \cite{2016msr-vtt} datasets.
Our TRIP is trained over the training set of WebVid-10M, and the validation sets of all three datasets are used for evaluation.
The \textbf{WebVid-10M} dataset consists of about $10.7M$ video-caption pairs with a total of $57K$ video hours.
$5K$ videos are used for validation and the remaining videos are utilized for training.
We further sample $2,048$ videos from the validation set of WebVid-10M for evaluation.
\textbf{DTDB} contains more than $10K$ dynamic texture videos of natural scenes.
We follow the standard protocols in \cite{2021stochastic} to evaluate models on a video subset derived from 4 categories of scenes (i.e., fires, clouds, vegetation, and waterfall).
The texts of each category are used as the text prompt since there is no text description.
For the \textbf{MSR-VTT} dataset, there are $10K$ clips in training set and $2,990$ clips in validation set.
Each video clip is annotated with $20$ text descriptions.
We adopt the standard setting in \cite{2023modelscope} to pre-process the video clips in validation set for evaluation.

\textbf{Implementation Details.}
We implement our TRIP on PyTorch platform with Diffusers \cite{Diffusers} library.
The 3D-UNet \cite{2023modelscope} derived from Stable-Diffusion v2.1 \cite{2022ldm} is exploited as the video backbone. 
Each training sample is a 16-frames clip and the sampling rate is $4$ fps.
We fix the resolution of each frame as 256$\times$256, which is centrally cropped from the resized video.
The noise scheduler $\{\beta_t\}_{t=1}^{T}$ is set as linear scheduler ($\beta_1=1\times 10^{-4}$ and $\beta_T=2\times 10^{-2}$).
We set the number of time steps $T$ as $1,000$ in model training.
The sampling strategy for video generation is DDIM \cite{2021ddimp} with $50$ steps.
The 3D-UNet and TNF module are trained with AdamW optimizer (learning rate: $2\times 10^{-6}$ for 3D-UNet and $2\times 10^{-4}$ for TNF module). 
All experiments are conducted on 8 NVIDIA A800 GPUs.

\textbf{Evaluation Metrics.}
Since there is no standard evaluation metric for I2V task, we choose the metrics of frame consistency (F-Consistency) which is commonly adopted in video editing \cite{2023tune,2023glv} and Frechet Video Distance (FVD) \cite{2018fvd} widely used in T2V generation \cite{2023videofusion,2023align} for the evaluation on WebVid-10M.
Note that the vision model used to measure F-Consistency is the CLIP ViT-L/14 \cite{2021clip}.
Specifically, we report the frame consistency of the first 4 frames (F-Consistency$_4$) and all the 16 frames (F-Consistency$_{all}$). 
For the evaluation on DTDB and MSR-VTT, we follow \cite{2021stochastic,2023modelscope} and report the frame-wise Frechet Inception Distance (FID) \cite{2017gans} and FVD performances.

\begin{table}[t]
	\centering
  \vspace{-0.1in}
	\caption{Performances of F-Consistency (F-Consistency$_4$: consistency among the first four frames, F-Consistency$_{all}$: consistency among all frames) and FVD on WebVid-10M.}
	\label{table:webvid_compare}
	\vspace{-5pt}
	\setlength{\tabcolsep}{0.3cm}
	\resizebox{\linewidth}{!}{
		\begin{tabular}{lccc}
			\toprule
			\textbf{Approach}     & \textbf{F-Consistency$_4$} ($\uparrow$) & \textbf{F-Consistency$_{all}$} ($\uparrow$)  & \textbf{FVD} ($\downarrow$)    \\
			\midrule
			T2V-Zero \cite{2023text2video-zero}
			& $91.59$ & $92.15$ & $279$ \\
			VideoComposer \cite{2023videocomposer}
			& $88.78$ & $92.52$ & $231$ \\ \midrule
			TRIP
			& \textbf{95.36} & \textbf{96.41} & \textbf{38.9} \\
			\bottomrule
	\end{tabular}}
 \vspace{-15pt}
\end{table}

\textbf{Human Evaluation.}
The user study is also designed to assess the temporal coherence, motion fidelity, and visual quality of the generated videos.
In particular, we randomly sample $256$ testing videos from the validation set in WebVid-10M for evaluation.
Through the Amazon MTurk platform, we invite $32$ evaluators, and each evaluator is asked to choose the better one from two synthetic videos generated by two different methods given the same inputs.

\subsection{Comparisons with State-of-the-Art Methods}
We compare our TRIP with several state-of-the-art I2V generation methods, e.g., VideoComposer \cite{2023videocomposer} and T2V-Zero \cite{2023text2video-zero}, on WebVid-10M. 
For the evaluation on DTDB, we set AL \cite{2019animating} and cINN \cite{2021stochastic} as baselines. 
Text-to-Video generation advances, e.g., CogVideo \cite{2022cogvideo} and Make-A-Video \cite{2022make-a-video} are included for comparison on MSR-VTT.

\begin{figure}
	\centering
 \vspace{-0.2in}
	\includegraphics[width=0.80\linewidth]{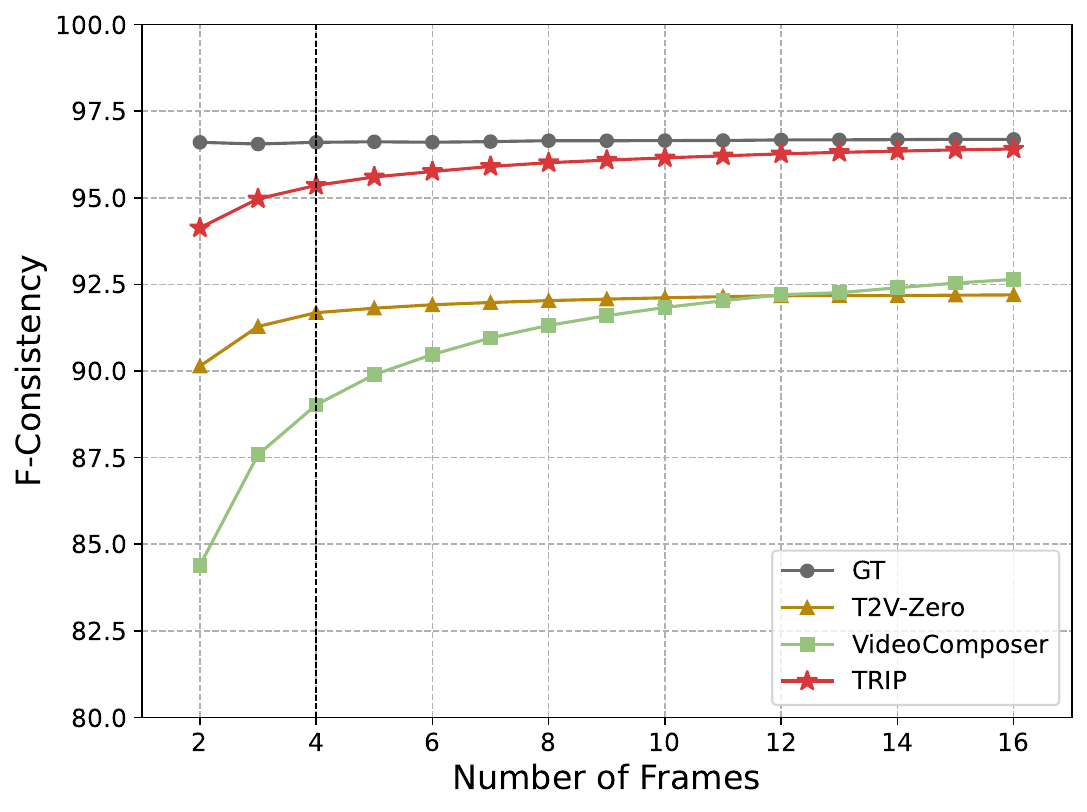}
	\vspace{-0.15in}
	\caption{Performance comparisons of F-Consistency by using different number of frames on WebVid-10M.}
	\label{fig:fc-index}
	\vspace{-0.15in}
\end{figure}

\begin{figure*}
	\centering
    \vspace{-0.2in}
	\includegraphics[width=0.85\linewidth]{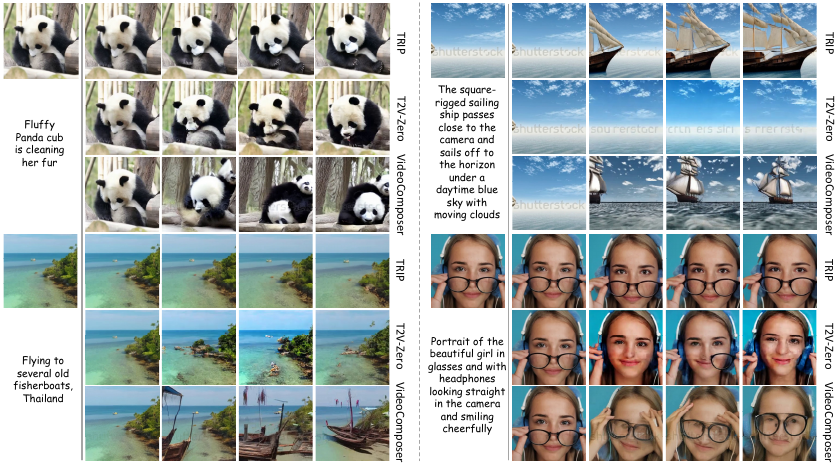}
	\vspace{-0.1in}
	\caption{Examples of I2V generation results by three methods (VideoComposer, T2V-Zero, and our TRIP) on WebVid-10M dataset. We uniformly sample four frames of each generated video for visualization.}
	\label{fig:webvid}
	\vspace{-0.2in}
\end{figure*}

\textbf{Evaluation on WebVid-10M.}
Table \ref{table:webvid_compare} summarizes performance comparison on WebVid-10M.
Overall, our TRIP consistently leads to performance boosts against existing diffusion-based baselines across different metrics.
Specifically, TRIP achieves the F-Consistency$_4$ of $95.36\%$, which outperforms the best competitor T2V-Zero by $3.77\%$.
The highest frame consistency among the first four frames of our TRIP generally validates the best temporal coherence within the beginning of the whole video. Such results basically demonstrate the advantage of exploring image noise prior as the guidance to strengthen the faithful visual alignment between given image and subsequent frames.
To further support our claim, Figure \ref{fig:fc-index} visualizes the frame consistency of each run which is measured among different numbers of frames. 
The temporal consistencies of TRIP across various frames constantly surpass other baselines.
It is worthy to note that TRIP even shows large superiority when only calculating the consistency between the first two frames. This observation again verifies the effectiveness of our residual-like noise prediction paradigm that nicely preserves the alignment with given image.
Considering the metric of FVD, TRIP also achieves the best performance, which indicates that the holistic motion dynamics learnt by our TRIP are well-aligned with the ground-truth video data distribution in WebVid-10M.

\begin{table}[t]
	\centering
	\caption{Performances of averaged FID and FVD over four scene categories on DTDB dataset.}
	\label{table:dtdb_compare}
	\vspace{-5pt}
	\setlength{\tabcolsep}{0.7cm}
	\resizebox{\linewidth}{!}{
		\begin{tabular}{lccc}
			\toprule
			\textbf{Approach}   & \textbf{Zero-shot} & \textbf{FID} ($\downarrow$)  & \textbf{FVD} ($\downarrow$)    \\
			\midrule
			AL~\cite{2019animating}
			& No & $65.1$ & $934.2$ \\
			cINN~\cite{2021stochastic}
			& No & $31.9$ & $451.6$ \\ \midrule
			TRIP
			& Yes & \textbf{24.8} & \textbf{433.9} \\
			\bottomrule
	\end{tabular}}
	\vspace{-0.2in}
\end{table}

Figure \ref{fig:webvid} showcases four I2V generation results across three methods (VideoComposer, T2V-Zero, and our TRIP) based on the same given first frame and text prompt.
Generally, compared to the two baselines, our TRIP synthesizes videos with higher quality by nicely preserving the faithful alignment with the given image \& text prompt and meanwhile reflecting temporal coherence. For example, T2V-Zero produces frames that are inconsistent with the second prompt ``the ship passes close to the camera'' and VideoComposer generates temporally inconsistent frames for the last prompt ``the portrait of the girl in glasses''. In contrast, the synthesized videos of our TRIP are more temporally coherent and better aligned with inputs.   
This again confirms the merit of temporal residual learning with amplified guidance of image noise prior for I2V generation.

\begin{figure}
	\centering
	\includegraphics[width=0.9\linewidth]{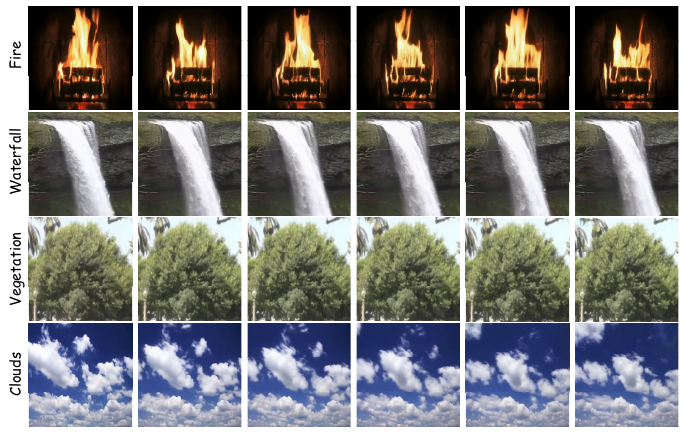}
	\vspace{-0.1in}
	\caption{Image-to-video generation results of four different natural scenes by our TRIP on DTDB dataset.}
	\label{fig:DTDB}
	\vspace{-0.2in}
\end{figure}

\textbf{Evaluation on DTDB.}
Next, we evaluate the generalization ability of our TRIP on DTDB dataset in a zero-shot manner.
Table \ref{table:dtdb_compare} lists the performances of averaged FID and FVD over four scene-related categories.
In general, our TRIP outperforms two strong stochastic models (AL and cINN) over both two metrics.
Note that cINN also exploits the residual information for I2V generation, but it combines image and residual features as clip feature by learning a bijective mapping. Our TRIP is fundamentally different in that we integrate residual-like noise estimation into the video denoising procedure to ease temporal modeling in the diffusion model.
Figure \ref{fig:DTDB} shows the I2V generation results of four different scenes produced by our TRIP. 
As shown in this figure, TRIP manages to be nicely generalized to scene-related I2V generation task and synthesizes realistic videos conditioned on the given first frame plus class label.

\textbf{Evaluation on MSR-VTT.}
We further compare our TRIP with recent video generation advances on MSR-VTT. Note that here we also evaluate TRIP in a zero-shot manner. Table \ref{table:MSR-VTT_compare} details the frame-wise FID and FVD of different runs. 
Similar to the observation on DTDB, our TRIP again surpasses all baselines over both metrics. The results basically validate the strong generalization ability of our proposal to formulate the open-domain motion dynamics.

\begin{table}[t]
	\centering
  \vspace{-0.2in}
	\caption{Performances of FID and FVD on MSR-VTT dataset.}
	\label{table:MSR-VTT_compare}
	\vspace{-5pt}
	\setlength{\tabcolsep}{0.8cm}
	\resizebox{\linewidth}{!}{
		\begin{tabular}{lccc}
			\toprule
			\textbf{Approach} & \textbf{Model Type}  & \textbf{FID} ($\downarrow$)  & \textbf{FVD} ($\downarrow$)    \\
			\midrule
			CogVideo~\cite{2022cogvideo} & T2V
			& $23.59$ & $1294$ \\
			Make-A-Video~\cite{2022make-a-video} & T2V
			& $13.17$ & - \\ 
			VideoComposer~\cite{2023videocomposer} & T2V
			& - & $580$ \\
			ModelScopeT2V~\cite{2023modelscope} & T2V
			& $11.09$ & $550$ \\ \midrule \midrule
			VideoComposer~\cite{2023videocomposer} & I2V
			& $31.29$ & $208$ \\ 
			TRIP & I2V
			& $\textbf{9.68}$ & $\textbf{91.3}$ \\
			\bottomrule
	\end{tabular}}
	\vspace{0.0in}
\end{table}

\begin{table}[t]
	\centering
	\caption{Human evaluation of the preference ratios between TRIP v.s. other approaches on WebVid-10M dataset.}
	\label{table:human_eval}
	\vspace{-5pt}
	\setlength{\tabcolsep}{0.4cm}
	\resizebox{\linewidth}{!}{
		\begin{tabular}{lcc}
			\toprule
			\textbf{Evaluation Items} & \textbf{v.s. T2V-Zero} & \textbf{v.s. VideoComposer}     \\
			\midrule
			Temporal Coherence
			& \textbf{96.9} v.s. 3.1 &  \textbf{84.4} v.s. 15.6 \\
			Motion Fidelity
			& \textbf{93.8} v.s. 6.2 &  \textbf{81.3} v.s. 18.7  \\
			Visual Quality
			& \textbf{90.6} v.s. 9.4 &  \textbf{87.5} v.s. 12.5  \\
			\bottomrule
	\end{tabular}}
	\vspace{-0.2in}
\end{table}

\subsection{Human Evaluation}
In addition to the evaluation over automatic metrics, we also perform human evaluation to investigate user preference with regard to three perspectives (i.e., temporal coherence, motion fidelity, and visual quality) across different I2V approaches. 
Table \ref{table:human_eval} shows the comparisons of user preference on the generated videos over WebVid-10M dataset.
Overall, our TRIP is clearly the winner in terms of all three criteria compared to both baselines (VideoComposer and T2V-Zero).
This demonstrates the powerful temporal modeling of motion dynamics through our residual-like noise prediction with image noise prior.

\subsection{Ablation Study on TRIP}
In this section, we perform ablation study to delve into the design of TRIP for I2V generation.
Here all experiments are conducted on WebVid-10M for performance comparison.

\textbf{First Frame Condition.}
We first investigate different condition strategies to exploit the given first frame as additional condition for I2V generation.
Table \ref{table:first} summarizes the performance comparisons among different variants of our TRIP.
\textbf{TRIP$_C$} follows the common recipe \cite{2023videocomposer} to concatenate the clean image latent code of first frame with the noise video latent code along channel dimension.
\textbf{TRIP$_{TE}$} is implemented by concatenating image latent code at the end of the noise video latent code along temporal dimension, while our proposal (\textbf{TRIP}) temporally concatenates image latent code at the beginning of video latent code.
In particular, TRIP$_{TE}$ exhibits better performances than TRIP$_{C}$, which indicates that temporal concatenation is a practical choice for augmenting additional condition in I2V generation.
TRIP further outperforms TRIP$_{TE}$ across both F-Consistency and FVD metrics.
Compared to concatenating image latent code at the end of video, augmenting video latent code with input first frame latent code at the start position basically matches with the normal temporal direction, thereby facilitating temporal modeling.

\begin{table}[t]
	\centering
	\caption{Performance comparisons among different condition approaches on WebVid-10M dataset.}
	\label{table:first}
	\vspace{-5pt}
	\setlength{\tabcolsep}{0.3cm}
	\resizebox{\linewidth}{!}{
		\begin{tabular}{lccc}
			\toprule
			\textbf{Model}     & \textbf{F-Consistency$_4$} ($\uparrow$) & \textbf{F-Consistency$_{all}$} ($\uparrow$)  & \textbf{FVD} ($\downarrow$)    \\
			\midrule
			TRIP$_C$
			& $94.77$ & $96.13$ & $41.3$ \\
			TRIP$_{TE}$
			& $95.17$ & $96.20$ & $39.8$ \\ \midrule
			TRIP
			& $\textbf{95.36}$ & $\textbf{96.41}$ & $\textbf{38.9}$ \\
			\bottomrule
	\end{tabular}}
	\vspace{-0.1in}
\end{table}

\begin{table}[t]
	\centering
	\caption{Evaluation of temporal residual learning in terms of F-Consistency and FVD on WebVid-10M dataset.}
	\label{table:trip}
	\vspace{-5pt}
	\setlength{\tabcolsep}{0.3cm}
	\resizebox{\linewidth}{!}{
		\begin{tabular}{lccc}
			\toprule
			\textbf{Model}     & \textbf{F-Consistency$_4$} ($\uparrow$) & \textbf{F-Consistency$_{all}$} ($\uparrow$)  & \textbf{FVD} ($\downarrow$)    \\
			\midrule
			TRIP$^-$
			& $94.66$ & $95.92$ & $39.9$ \\
			TRIP$^W$
			& $95.22$ & $95.96$ & $43.0$ \\ \midrule
			TRIP
			& $\textbf{95.36}$ & $\textbf{96.41}$ & $\textbf{38.9}$ \\
			\bottomrule
	\end{tabular}}
	\vspace{-0.25in}
\end{table}

\begin{figure*}
	\centering
	\includegraphics[width=0.85\linewidth]{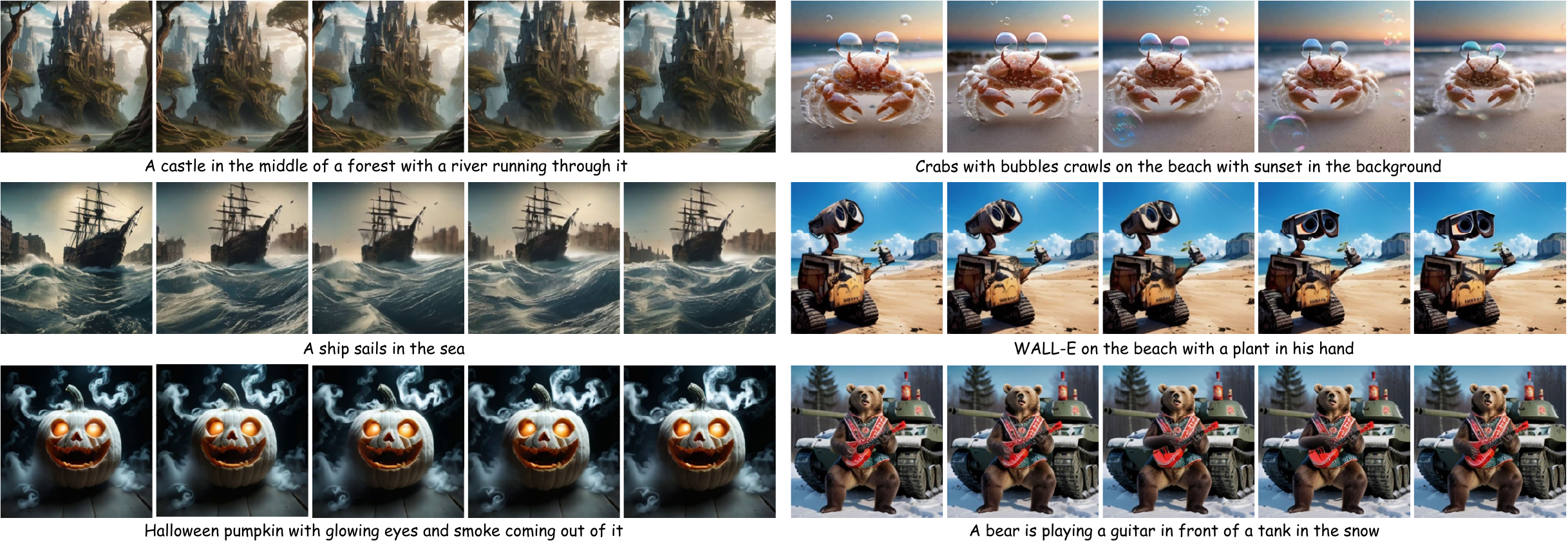}
	\vspace{-0.1in}
	\caption{Customized image animation with Stable-Diffusion XL \cite{2023SDXL} for text-to-video generation. We first employ Stable-Diffusion XL to synthesize fancy images and then exploit TRIP for image animation. Six generated videos with corresponding text prompts are presented.}
	\label{fig:T2I2V}
	\vspace{-0.2in}
\end{figure*}

\begin{figure}
	\centering
	\includegraphics[width=0.95\linewidth]{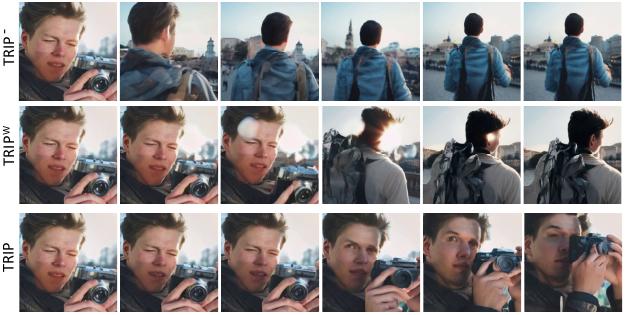}
	\vspace{-0.1in}
	\caption{Visualization of I2V example with text prompt ``Young male tourist taking a photograph of old city on sunset'' by using variants of our TRIP.}
	\label{fig:ablation-visual}
	\vspace{-0.2in}
\end{figure}

\textbf{Temporal Residual Learning.}
Next, we study how each design of our TRIP influences the performances of both temporal consistency and visual quality for I2V generation.
Here we include two additional variants of TRIP: (1) \textbf{TRIP}$^-$ removes the shortcut path of temporal residual learning and only capitalizes on 3D-UNet for noise prediction; (2) \textbf{TRIP}$^W$ is a degraded variant of temporal residual learning with image noise prior by simply fusing reference and residual noise via linear fusion strategy.
As shown in Table \ref{table:trip}, {TRIP}$^W$ exhibits better performances than {TRIP}$^-$ in terms of F-consistency.
The results highlight the advantage of leveraging image noise prior as the reference to amplify the alignment between the first frame and subsequent frames.
By further upgrading temporal residual learning (\textbf{TRIP}$^W$) with Transformer-based temporal noise fusion, our TRIP achieves the best performances across all metrics.
Figure \ref{fig:ablation-visual} further showcases the results of different variants, which clearly confirm the effectiveness of our designs.

\subsection{Application: Customized Image Animation}

\begin{figure}
	\centering
	\includegraphics[width=0.93\linewidth]{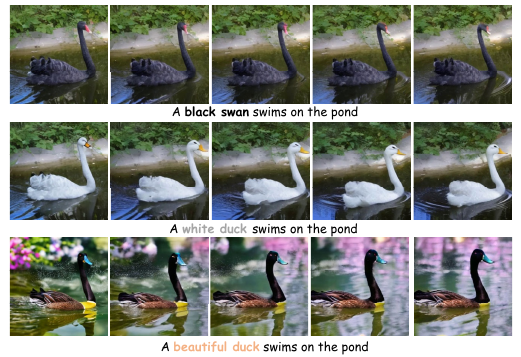}
	\vspace{-0.1in}
	\caption{Customized image animation with image editing models (e.g., InstructPix2Pix \cite{2023instructpix2pix} and ControlNet \cite{2023adding}). The first row shows image animation results given the primary real-world first image. The second and third rows show I2V results given the edited images via InstructPix2Pix and ControlNet, respectively.}
	\label{fig:T2I2V-edit}
	\vspace{-0.2in}
\end{figure}

In this section, we provide two interesting applications of customized image animation by directly extending TRIP in a tuning-free manner.
The first application is text-to-video pipeline that first leverages text-to-image diffusion models (e.g., Stable-Diffusion XL \cite{2023SDXL}) to synthesize fancy images and then employs our TRIP to animate them as videos.
Figure \ref{fig:T2I2V} shows six text-to-video generation examples by using Stable-Diffusion XL and TRIP.
The results generally demonstrate the generalization ability of our TRIP when taking text-to-image results as additional reference for text-to-video generation.
To further improve controllability of text-to-video generation in real-world scenarios, we can additionally utilize image editing models (e.g., InstructPix2Pix \cite{2023instructpix2pix} and ControlNet \cite{2023adding}) to modify the visual content of input images and then feed them into TRIP for video generation.
As shown in Figure \ref{fig:T2I2V-edit}, our TRIP generates vivid videos based on the edited images with promising temporal coherence and visual quality.

\section{Conclusions}
This paper explores inherent inter-frame correlation in diffusion model for image-to-video generation.
Particularly, we study the problem from a novel viewpoint of leveraging temporal residual learning with reference to the image noise prior of given image for enabling coherent temporal modeling.
To materialize our idea, we have devised TRIP, which executes image-conditioned noise prediction in a residual fashion.
One shortcut path is novelly designed to learn image noise prior based on given image and noised video latent code, leading to reference noise for enhancing alignment between synthesized frames and given image.
Meanwhile, another residual path is exploited to estimate the residual noise of each frame through eased temporal modeling.
Experiments conducted on three datasets validate the superiority of our proposal over state-of-the-art approaches.

{
 \small
 \bibliographystyle{ieeenat_fullname}
 \bibliography{main}
}

\end{document}